\documentclass[twoside,leqno,twocolumn]{article}

\usepackage[letterpaper]{geometry}

\usepackage{ltexpprt}
\usepackage{xcolor}
\usepackage{amssymb}
\usepackage{graphicx}
\usepackage{amsmath}
\usepackage[T1]{fontenc}
\begin{document}

\title{\Large Estimator Vectors: OOV Word Embeddings based on Subword and Context Clue Estimates}
\author{Raj Patel \thanks{rpatel17@masonlive.gmu.edu, George Mason University, Fairfax, VA.}
\and Carlotta Domeniconi \thanks{cdomenic@gmu.edu, George Mason University, Fairfax, VA.}}

\date{}

\maketitle


\fancyfoot[R]{\scriptsize{Copyright \textcopyright\ 20XX by SIAM\\
Unauthorized reproduction of this article is prohibited}}





\begin{abstract} \small\baselineskip=9pt Semantic representations of words have been successfully extracted from unlabeled corpuses using neural network models like word2vec. These representations are generally high quality and are computationally inexpensive to train, making them popular. However, these approaches generally fail to approximate out of vocabulary (OOV) words, a task humans can do quite easily, using word roots and context clues. This paper proposes a neural network model that learns high quality word representations, subword representations, and context clue representations jointly.  Learning all three types of representations together enhances the learning of each, leading to enriched word vectors, along with strong estimates for OOV words, via the combination of the corresponding context clue and subword embeddings.  Our model, called Estimator Vectors (EV), learns strong word embeddings and is competitive with state of the art methods for OOV estimation.\end{abstract}

\section{Introduction}
Semantic representations of words are useful for many natural language processing (NLP) tasks.  While there exists many ways to learn them, models like word2vec \cite{mikolov2013efficient} and GloVe \cite{pennington2014glove} have been shown to be very efficient at producing high quality word embeddings.  These embeddings not only capture similarity between words, but also capture some algebraic relationships between words.  These models, though, also have some downsides.  One major drawback is that they can only learn  embeddings for words in the vocabulary, determined by the corpus they were trained on.  Although common words are typically captured, most existing approaches are unable to learn the meaning of new words, known as out of vocabulary (OOV) words, a task humans can do easily.  Unknown words could be new words or domain specific words, both of which could be very important for NLP tasks.  Therefore, finding good representations for these words poses a relevant challenge.  Some attempts have been made to estimate representations of OOV words, generally based on how humans learn a new word.  One way is to use external auxiliary information, like definitions of the word \cite{bahdanau2017learning}. One downside of this is that it requires external information, which may not be accessible.  Another way that gets around this problem is to estimate OOV word representations using word roots or subwords \cite{bojanowski2017enriching}, \cite{pinter2017mimicking}.  This approach can work well, but struggles on words that have less meaningful word roots.  Another strategy is to use the context the OOV word appears in (in human learning, these are known as {\textit{context clues}}).  These methods estimate OOV representations by adding the context word representations \cite{lazaridou2017multimodal}, \cite{horn2017context}, or by training the representation with these words \cite{herbelot2017high}.  Context words are generally good for estimating an unknown word, but these methods can struggle with weighing the important context clues over the less important ones.

In this paper, we propose Estimator Vectors (EV), a new neural network approach that learns three types of embeddings: word embeddings, context clue embeddings, and subword embeddings.  The word embeddings are similar to other word embedding methods, while the context clue and subword embeddings are used to estimate word embeddings when they are encountered.  This approach learns these embeddings jointly, enhancing the quality of each.

The major contribution of this work is  a novel and effective approach (Estimator Vectors, or EV) to word embedding and out-of-vocabulary estimation with the following distinctive features:
(1) EV learns and uses three sets of vectors, context clue embeddings, subword embeddings,  {\it{and}} word embeddings, each for its own specific purpose;
(2) EV learns these embeddings at the same time, in order to learn the three sets of vectors effectively.  EV learns context clue embeddings and subword embeddings such that their individual averages estimates the representation of the target word. At the same time, it uses the target estimate in the target context pair to learn word embeddings.  The interplay of the three embeddings  enhances the learning of all three, leading to strong estimates.

The rest of the paper is organized as follows. Section \ref{relatedwork} discusses the background and related work. Section \ref{estimator_vectors} defines  EV in detail.  Section \ref{experiments} describes the experimental setup, and Section \ref{results} discusses the results.  Finally, Section \ref{conclusion} concludes the paper.

\section{Related Work}
\label{relatedwork}
In this section we discuss relevant previous work.  First, we discuss word embeddings in general, then focus on strategies for estimating OOV words.
\subsection{Word Embeddings}
Word2vec \cite{mikolov2013efficient}  is a popular approach for computing  semantic representations of words.  It relies on training a shallow neural network model based on predicting the context of words, using backpropagation.  The original model had two versions; continuous bag of words, which uses the context of a word to predict the word itself, and skipgram, which uses the word to predict its context \cite{mikolov2013efficient}.  Both of these models input a 1-hot encoding of each word, and output a softmax probability distribution over the vocabulary.  The word vectors are the first layer weights connected to a word's index.  Due to the large amount of calculations needed to compute the softmax distribution over the entire vocabulary (usually a very large number), Mikolov et al. \cite{mikolov2013distributed} proposed a more efficient method, known as negative sampling.  Instead of calculating the probability of every word in the vocabulary, negative sampling tries to maximize the probability of a target word co-occurring with its context words, while also minimizing the probability of the target word co-occurring with randomly selected words, known as negative samples.  In the negative sampling version of the skipgram model, the probability of a target word $w_{t}$ and a context word $w_{c}$ occurring together is calculated as:
\begin{equation}\label{skipgram_eq}
\sigma (u_{w_{t}}\cdot v_{w_{c}})
\end{equation}
where $\sigma(x) = 1/({1 + e^{-x}})$ is the logistic function,
$u_{w_{t}}$ is the word vector representation for the target word ${w_{t}}$, and $v_{w_{c}}$ is the word vector representation for the context word ${w_{c}}$.  In this formulation, target context pairs with similar representations will have high dot products, leading to a high value (closer to 1), while different representations will have a low dot product, leading to low values (closer to 0).  The overall loss function for negative sampling with the skipgram model is the negative log likelihood:
\begin{equation}
E =   -log  \sigma (u_{w_{t}}\cdot v_{w_{c}})  -  \sum_{n\in N} log  \sigma (-u_{w_{t}}\cdot v_{n}) 
\end{equation}
where $N$ is the set of negative samples.  

To demonstrate how the skipgram model is trained, we use the example sentence ``The yellow car sped up quickly''.  If the target word ${w_{t}}$ was ``car'',  one context word ${w_{c}}$ could be ``sped''.  Skipgram learns a representation such that $\sigma (u_{car}\cdot v_{sped})$ is a large value, while the probability of co-occurrence with negative samples, like $\sigma (u_{car}\cdot v_{coffee})$, is  small.

Continuous bag of words also has a negative sampling version \cite{mikolov2013distributed}.  Continuous bag of words pairs the sum of the context vectors with the target vector, and learns a probability that is high when the sum of the context is paired with its target and low when paired with negative samples.  The probability is calculated as:
\begin{equation}
\sigma (\sum_{c\in C} v_{w_{c}}\cdot u_{w_{t}})
\label{cbow-eq}
\end{equation}
where $C$ is the set of all the context words for the target.  

Continuous bag of words trains all the context words at once.  For the example ``The yellow car sped up quickly'', continuous bag of words learns representations such that  $\sigma ((v_{the}+v_{yellow}+v_{sped}+v_{up})\cdot u_{car})$ is a large value, while with a negative sample like ``coffee'', $\sigma ((v_{the}+v_{yellow}+v_{sped}+v_{up})\cdot u_{coffee})$ is a small value.

Both the skipgram model and the continuous bag of words model learn embeddings on a subsampled version of the training corpus, in order to reduce the influence of overly frequent words.  Mikolov et al. \cite{mikolov2013distributed} found that removing some instances of very frequent words improves the general quality of the word embeddings.  Therefore, word instances are removed with a probability based on their frequency, where more frequent words have a higher probability of being removed.

\subsection{Out-Of-Vocabulary Embeddings}
Models like word2vec lead to effective word representations, but only for words in the vocabulary of the original training corpus.  Therefore, models using word2vec representations struggle when encountering OOV words.  There have been attempts to estimate OOV words' vector representations.  These approaches tend to mirror human strategies for learning new words; by using the word's roots/subwords or using the context the word was found in. 

\subsubsection{Subword Based Approaches}
  One  way to estimate an OOV word's embedding is to use its subword information.  Bojanowski et al. \cite{bojanowski2017enriching} train both word vectors and subword vectors at the same time, such that the sum of subwords' vectors approximates the target word.  Their model, known as fastText, uses a similar approach to the negative sampling skipgram model, but replaces the target representation $u_{w_{t}}$ with the sum of its $n$-gram vectors, so the probability (of $w_{t}$ and $w_{c}$ co-occuring) becomes:
\begin{equation}
\sigma (\sum_{g\in G_{w_{t}}}z_{g}\cdot v_{w_{c}})
\end{equation}
where $G_{w_{t}}$ is the set of character $n$-grams (the subwords) of the target word $w_{t}$, and $z$ is the embedding of the subwords.

The subword model is similar to skipgram, except it estimates the target word's embedding with the sum of its subword vectors.  In the recurring example, if $n=2$ for the character $n$-grams, the probability $\sigma (u_{car}\cdot v_{sped})$ of skipgram is replaced with $\sigma ((z_{<c}+z_{ca}+z_{ar}+z_{r>}+z_{<car>})\cdot v_{sped})$, where the ``\textless'' and ``\textgreater'' denote the beginning and end of a word respectively,  and ``\textless car \textgreater'' is a special token added to the subword set.

Subword methods are powerful, but do have some flaws.  They struggle with words that have weak or unknown word roots.  For example, subword methods may struggle with foreign words, as they may not learn the subwords required for a good estimate.

\subsubsection{Context Based Approaches}
Other methods use the context of an OOV word to estimate its representation.  Some methods simply sum the existing context word embeddings to get an estimate of the OOV word  \cite{lazaridou2017multimodal}, \cite{horn2017context}.  This summation method works due to the algebraic property of word2vec, which states that simple algebraic operations can be applied to word vectors in order to mimic the operation semantically (for example, $u$(``King'') - $u$(``Man'') + $u$(``Woman'') $\approx$ $u$(``Queen'')) \cite{mikolov2013efficient}.  This suggests that adding relevant words like the context words should give a good estimate of the OOV word representation, which is demonstrated in \cite{lazaridou2017multimodal}, \cite{horn2017context}.  Other OOV estimation methods refine the estimate using various techniques.  Nonce2vec \cite{herbelot2017high} trains the estimate in the existing skipgram model using a very high learning rate. Another technique, \textit{\`{a} la carte}, \cite{khodak2018carte} refines the estimate with a linear transformation learned from the original training corpus and set of word vectors.

The above models do have limitations when it comes to estimating OOV embeddings.  They tend to be based on the summation of their word embeddings, which may have weaknesses, depending on the original embedding method. The word2vec models only focus on learning word embeddings, which may hinder their ability to estimate OOV words.  The skipgram model learns with one target context pair at a time, and therefore is not capable of learning how multiple context words relate to each other in terms of impact, nor what each context word says about the target word relative to the others.  This means when the context words are summed to estimate an OOV word embedding, the estimate will be less accurate, as each words' relative importance (how much it affects the summed vector)  and relative contribution (what it adds when being summed with other context words) are not captured by a skipgram model.  The continuous bag of words model, on the other hand, does learn how word vectors relate to each other by summing the context words, as it is trained on this sum. However, it does not learn infrequent words' embeddings well \cite{naili2017comparative}.  This is because it learns all of its vectors at once (the sum of the context vectors, being paired with the target), diluting infrequent words (as the model does not learn infrequent words individually, decreasing how much it learns per target word).  This may interfere with how well the sum of context words can estimate an OOV word, as we expect less frequent words to be informative for an OOV word.

\subsubsection{Combined Approaches}
\label{combined_approaches}
The Form-Context model \cite{schick2019learning} combines subwords and context words to estimate OOV words.  This approach takes previously trained word embeddings and a text corpus, and trains a model that can estimate an OOV word embedding using subword information and context information.  For the subword representation (which they call the form representation), it learns a set subword embedding for each character ngram, similar to fastText \cite{bojanowski2017enriching}.  For the context representation, it takes the average of the context words, and learns a linear transformation to estimate to OOV word, similar to \textit{\`{a} la carte} \cite{khodak2018carte} .  It then combines the embeddings in a weighted sum:

\begin{equation}
v = \alpha \cdot v_{context} +  (1 -\alpha) \cdot v_{form}
\end{equation}

where $\alpha$ decides how much to weigh the context estimate against the subword estimate.  Schick and Sch{\"u}tze \cite{schick2019learning} propose two different ways to calculate $\alpha$.  The first approach, known as single parameter, simply learns $\alpha$ as one value for every subword context pair.  The second approach, known as the gated model, learns parameters to calculate $\alpha$  with the following function:
\begin{equation}
\alpha  =   \sigma(w^T [v_{context}, v_{form}]+b)
\end{equation}

where $\sigma$ is the sigmoid function, and $w \in \mathbb{R}^{2k}$ and $b \in \mathbb{R}$ are learnable parameters ($k$ is the embedding dimensionality).  The gated model learns to weigh how important the context is compared to the subwords in each scenario.

The Form-Context model is very good at estimating OOV words, but does have some drawbacks.  First, it is dependant on the quality of word embeddings it is trained on, as it does not learn its own.  Because the subword embeddings and context linear transformation are trained to estimate the word embeddings directly, the word embeddings do not learn from subword or combined context information, which could enrich the quality of the embeddings.

\section{Estimator Vectors}
\label{estimator_vectors}
\subsection{Model}
We present Estimator Vectors (EV), a word2vec based model that learns three types of representations: word embeddings, context clue embeddings, and subword embeddings. EV can easily create an embedding for an OOV word as it is encountered.  Like the Form-Context model, Estimator Vectors combine both context and subword information.  However, EV has some key differences.  First, it learns word embeddings, subword embeddings, and context embeddings at the same time.  This joint training leads to stronger representations.  Second, unlike other methods' context estimations, EV learns a unique set of context vectors, which we call context clue embeddings.  Unlike word embeddings, these embeddings are trained in a sum, and therefore not only learn the meaning of each context word, but also how "informative" it is.  This means that more informative words will have a greater impact on the sum, leading to a strong context estimate.

Our model takes a similar approach as the skipgram \cite{mikolov2013efficient} and fastText \cite{bojanowski2017enriching} models mentioned above.  EV trains on word co-occurrence pairs, but replaces the first word embedding ($u_{w_t}$ in equation \ref{skipgram_eq}) with a context clue estimate, which is the average of the context clue embeddings for the context words of $w_t$:
\begin{equation}
cc_{w_t} = \frac{1}{|Q_{w_{t}}|}\sum_{q \in Q_{w_{t}}} h_{q}
\end{equation}

where $Q_{w_t}$ is the set of context clues for $w_t$ and $h$ is the set of context clue embeddings.

In addition, it replaces the first word with the subword estimate as well, which is the average of the character n-gram embeddings for $w_t$:
\begin{equation}
sub_{w_t} = \frac{1}{|G_{w_{t}}|}\sum_{g\in G_{w_{t}}}z_{g}
\end{equation}
where $G_{w_{t}}$ is the set of character $n$-grams (the subwords) of the target word $w_{t}$, and $z$ is the embedding of the subwords.

EV maximizes both probabilities for words that co-occur and minimizes both probabilities for the negative samples.  The equations for the context clue probability is
\begin{equation}
\sigma (cc_{w_t}\cdot  v_{w_{c}})
\end{equation}
where $v$ is the set of word embeddings. Similarly, the probability for the subwords is 
\begin{equation}
\sigma (sub_{w_t}\cdot v_{w_{c}})
\end{equation}

EV optimizes both probabilities at the same time, through the following error function:
\begin{multline} \label{eq:CCV}
E =   -log \sigma (cc_{w_t}\cdot  v_{w_{c}}) -log \sigma (sub_{w_t}\cdot v_{w_{c}})        \\
   - \sum_{n\in N} [ log  \sigma (- cc_{w_t}\cdot  v_{n}) + log  \sigma (- sub_{w_t}\cdot v_{n})]
\end{multline}
where $N$ is the set of negative samples. 

Note that two major components are being learned by this model; (1) the overall semantic space is being learned by the word vectors via the skipgram pairings and negative samples, and (2) the ability to estimate any word in the space is also being learned by the context clue and subword vectors.  In addition, since these are all being learned at the same time, they are enhanced by each other.  This means the word embeddings $v$ learn from both context clue embeddings $h$ and subword embeddings $z$, while $h$ and $z$ both learn from $v$.  In addition, since both impact $v$, $h$ indirectly learns from $z$ and vice versa.  This interplay between each type of vectors leads to high quality vectors of each type.

As an example, we return to the sentence ``The yellow car sped up quickly''.  Given the target context pair ``car'' and ``sped'', Estimator Vectors would train on the following probabilities:  $\sigma (\frac{1}{4}(h_{the}+h_{yellow}+h_{sped}+h_{up})\cdot v_{sped})$ and $\sigma (\frac{1}{5}(z_{<c}+z_{ca}+z_{ar}+z_{r>}+z_{<car>})\cdot v_{sped})$.  In this example, the EV model learns three things:  a semantic representation for ``sped'', how to estimate a semantic representation for ``car'' by learning representations for its context clues, and how to estimate ``car'' by learning representations for its subwords.  In addition, it learns all of these representations based on the fact that ``car'' and ``sped'' co-occur.

The learned word vectors $v$ can be used as normal word embeddings for downstream tasks.  When an OOV word $w_o$ is encountered, the context clue representation is calculated as:
\begin{equation}
cc(w_o) = \frac{1}{|Q_{w_{o}}|}\sum_{q \in Q_{w_{o}}} h_{q}
\end{equation}
and the subword representation is calculated as 
\begin{equation}
sub(w_o) = \frac{1}{|G_{w_{o}}|}\sum_{g\in G_{w_{o}}}z_{g}
\end{equation} .These estimates can then be combined for a final estimate of the OOV word:
\begin{equation}
est(w_o) = cc(w_o) + sub(w_o)
\end{equation}

\subsection{Postprocessing Context Clues}
\label{postprocessing}
One advantage of word embeddings is that they can be summed to estimate a new word.  However, sets of word vectors tend to share a few common directions, and summing multiple vectors can amplify these directions.  This can harm the sum's representation, as the uncommon directions tend to carry more meaning.  In order to reduce this problem, Mu and Viswanath \cite{mu2017all}  and Arora et al. \cite{arora2016simple} propose removing the top PCA components from the vectors.  

In order to improve the context clue representations, we remove the top three components based on the word representations from the sum of the context clues.  This is done before a context clue representation is combined with a subword representation.  We denote the postprocessed context clue representation as $cc'(w_o)$, which leads to the final equation:
\begin{equation}
est(w_o) = cc'(w_o) + sub(w_o)
\end{equation}

\section{Experiments}
\label{experiments}
\subsection{Baseline and Hyperparameters}
	We compare EV's word embeddings to the word2vec skipgram model and fastText. Both word2vec and fastText are trained using the gensim library, a very efficient embedding toolkit in Python \cite{rehurek2010software}.  All models, including EV, use the same hyperparameters: embeddings of size 300,  minimum frequency of 100, sampling with a rejection threshold (for reducing overly frequent words)  of .0001, window sizes between 1 and 5 (uniformly sampled), and 5 negative samples.  The models are trained for 5 epochs,  with a learning rate of 0.025, which linearly decays to a minimum of .0001, as training goes on.  For fastText, character ngrams from size 3 to 6 were selected.  For EV, the context clues of a target are taken from a window size of 3 (3 before and 3 after).  If a word with less than 100 frequency occurs as a context clue, it is ignored.  Additionally, the subwords used in EV were chosen as any ngram from size 3 to 5, and only those that occur in at least 3 words in the vocabulary (in order to compare to the Form-Context model, mentioned below).
	
    In addition to word embedding quality, we also show EV's ability to estimate OOV words effectively.  To this end, we compare to context based methods, subword methods, and combined methods.  For context models, EV is compared to simple summation methods of the skipgram vectors and  \textit{\`{a} la carte} embeddings \cite{khodak2018carte} (which are based on the trained word2vec embeddings).  We train \textit{\`{a} la carte} embeddings on the skipgram vectors mentioned above, with a minimum word count of 500 for training the linear transformation.  For subword methods, we compare to fastText \cite{bojanowski2017enriching}.  Finally, we compare to the state of the art combined method, the Form-Context model \cite{schick2019learning}.  We train the Form-Context model on the skipgram embeddings mentioned above, using the hyper parameters mentioned in \cite{schick2019learning}.  These include  a minimum word count of 100, and character ngrams from size 3 to 5 (ngrams only taken if they occur in at least 3 different words) . For weights, we examine the gated model, as it generally had stronger results than the single parameter model.

\subsection{Data Set}    
    All models were trained on the Westbury Wikipedia Corpus (WWC) \cite{shaoul2010westbury}.  We use a modified version provided by Khodak et al. \cite{khodak2018carte} where sentences with certain rarewords removed  for purpose of testing OOV estimation.  
    
    \subsection{Testing}
	The EV model trains word embeddings along with context clue embeddings and subword embeddings,  leading to two goals.  The first goal is for the word embeddings ($v$) trained by this model to be high quality.  To verify this, the trained embeddings are tested on an analogy task and a similarity to human judgement task.  The analogy task, first shown in \cite{mikolov2013efficient}, tests the embeddings on how well they can solve an analogy, like the $u$(``King'') - $u$(``Man'') + $u$(``Woman'') $\approx$ $u$(``Queen'') example mentioned earlier.  Three words (the left side of the equation) are used to estimate a new vector.  Then, this vector is compared to all word embeddings, and the most similar (by cosine similarity) is chosen.  The score is the percentage of correct words found.  The task is split into two parts: semantic (which captures meaningful relationships) and syntactic (which captures structural relationships).   The quality of the word embeddings is also evaluated using the WS353 task, created by Finkelstein et al. \cite{finkelstein2001placing}.  This task contains 353 word pairs with human created similarity scores for each pair.  These scores are compared to the cosine similarity between the corresponding word embeddings, using Spearman's rank order correlation coefficient \cite{spearman1904proof}.  Better embeddings should have a higher correlation coefficient.  For both the analogy and WS353 tasks, any analogy or pair involving words not in the vocabulary is ignored.

	The second goal of EV is for the context clue embeddings ($h$) and subword embeddings ($z$) to find good estimates of OOV word embeddings.  This is evaluated by two tasks for OOV estimation;  the definitional nonce task, created by Herbelot and Baroni \cite{herbelot2017high}  and the Contextualized Rare Word (CRW) task, created by Khodak et al. \cite{khodak2018carte}.  The definitional nonce task contains 300 sentences, each being the first sentence of the Wikipedia page for a nonce word.  The goal of this test is to pretend the nonce word is unknown, estimate it using the sentence, and then compare the estimated embedding to its original embedding.  Because the sentences are definitions, their contexts are known to be informative.  Note that the definitional nonce task compares the nonce estimate to its real location, and therefore must have a real embedding for the nonce.  All models trained from WWC are missing 5 nonce words, and therefore only evaluate based on 295 nonces. The second test is the CRW set. The goal of CRW is to estimate OOV words given the word and a set of contexts.  It is built on the Rare Word (RW) dataset \cite{luong2013better}, which has a list of rare words, pairs them with other words, and contains similarity scores for the pair based on human judgements.  The goal is to try to estimate the rare words such that their similarity to the paired word correlates with the human scores.  CRW extends this, by adding sets of contexts to each rare word.  CRW estimates the rare word with different amounts of contexts and judges how well their pair similarity correlates with human judgements.  Unlike the definitional nonce task, the CRW task does not require the words to have existing embeddings.
	
	\section{Results}
\label{results}

     Each result we present is the average of 10 trained versions of the corresponding model.  Statistical significance is assessed using a one-way ANOVA with a post-hoc Tukey HSD test with a p-value threshold equal to 0.05. For each task, boldface indicates the technique with the statistically significant best performance score.
     
     For the results, we denote the whole Estimator Vectors model as EV, with its word vectors as EV-word, its subword vectors as EV-s, and its context clue vectors as EV-c.  We denote the Form-Context model as FCM, and similarly denote the subword and context only models as FCM-s and FCM-c respectively.  In addition, skipgram is denoted as sg.

	\subsection{Word Embedding Quality}
	We compare gensim implementations of skipgram and fastText to the word vectors trained by our model.  Note that both FCM and \textit{\`{a} la carte} embeddings are based on the skipgram embeddings, so its word embedding quality is particularly important. The results for the analogy test and the WS353 task are shown in Table \ref{tab:analogyandws353}.

		\begin{table}[t]
		
		\begin{center} 

 		\begin{tabular}{@{}lr@{}lr@{}lr@{}l@{}} 
 		     & Semantic && Syntactic  &&  WS353 ($\rho$) &\\  
 		\hline
 		sg & 80.30!\% && 73.03\% && .7154 & \\ 
 		
 		fastText & 79.60\% && \textbf{75.97}\% && .7125  &\\
 		
 		EV-word & \textbf{85.30}\% && 67.42\% && \textbf{.7346} & \\ 
 		
 		\hline
		\end{tabular}

		 \end{center} 
        \caption{Analogy Task and WS353 Task Judgement}\label{tab:analogyandws353}
		\end{table}

These results show that EV word vectors are stronger embeddings, due to being trained jointly with subwords and context clues.  EV outperforms skipgram and fastText in the Semantic analogy test, along with the WS353 semantic similarity task.  However, it performs worse on the Syntactic task. This shows that joint training with subwords and context clues at least enhances the semantic information contained embeddings, although it may slightly decrease the structural information.  This may be mitigated by adding the EV-s vectors, as subword vectors tend to be good at capturing syntactic information \cite{bojanowski2017enriching}. This is shown by fastText,a subword model, which has the strongest Syntactic task score.

\subsection{OOV Embedding Estimation}
\begin{figure}[htp]
    \centering
    \includegraphics[width=8cm]{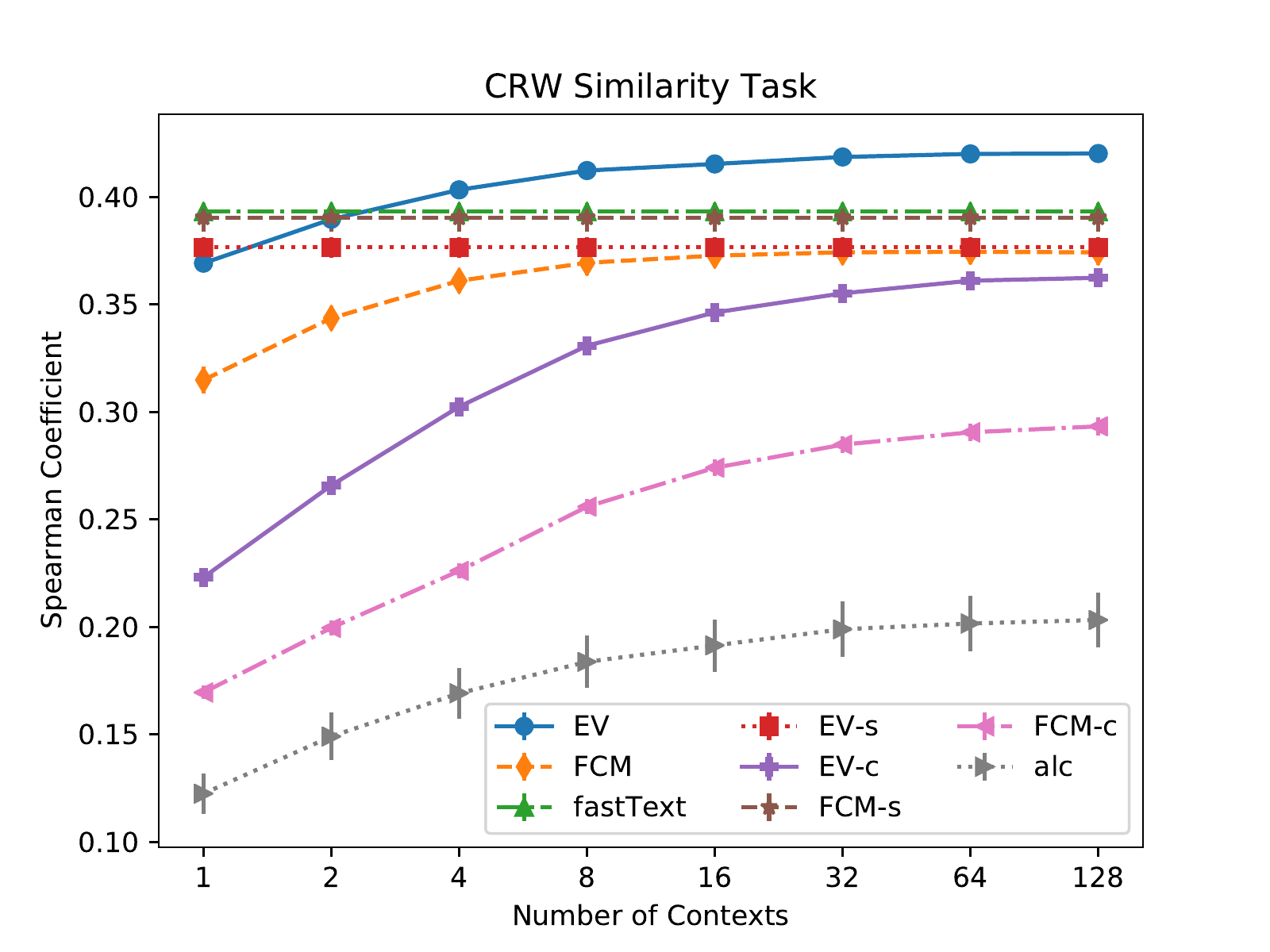}
    \caption{CRW Task}
    \label{fig:crw_graph}
\end{figure}

    We also investigate how well EV perform at OOV estimation.  We test on the definitional nonce task and the CRW task.  We show the results for the definition nonce task in Table \ref{tab:deftask}, and for the CRW task in Figure \ref{fig:crw_graph}.  The definitional nonce task measures the rank of the `real' embedding of the OOV word in the list of nearest neighbors of the estimated embedding. The ranks of each nonce are aggregated using two metrics: MRR and Median Rank.  MRR is the Mean Reciprocal Rank, i.e. the average inverse rank, and  higher values are better.  Since Median Rank measures rank, lower is better (with the vocabulary size 145741 being the worst rank).  For the CRW task, each method's correlation with human scores is shown across multiple context sizes, to show how they perform in both low and high context settings.
    
    We compare EV to the context based, subword based, and combined methods mentioned earlier. Note that fastText is incompatible with the definitional nonce task as a subword estimation method.  This is because fastText's word embeddings are already the sum of its subwords.  Therefore, its "real" embedding it the same as its subword estimation, which makes it unable to be judged by the definitional nonce task.  As a result, fastText is omitted from the definitional nonce task.

        \begin{table}[t]

		\begin{center} 
		 
 		\begin{tabular}{@{}lr@{}lr@{}l@{}} 
 		     & MRR && Median Rank&  \\ 
 		\hline
 		sg (additive) & .0264 && 265.8 &  \\ 
 		
 		\textit{\`{a} la carte} & .0481 && 246.7 &  \\
 		
		FCM-c & .0817 && 90.7 &  \\ 
		
		FCM-s & \textbf{.9327} && \textbf{1} &  \\ 
		
		FCM & .7426 && \textbf{1} &  \\ 
 		
		 EV-c & .0694 && 74.8 & \\ 
 		
 		 EV-s  & .8616 && \textbf{1} & \\ 
 		
 		EV & .6916 && \textbf{1} &  \\

 		\hline
		\end{tabular}
		 \end{center} 
         \caption{Definitional Nonce Task}\label{tab:deftask}
		\end{table}
        
The results for the definitional task (shown in Table \ref{tab:deftask}) show that EV performs strongly, although it is not the best model for the task. For the combined subword/context methods (which ideally should perform the best), the Form-Context model outperforms everything, including EV, in MRR.  For median rank, they both get the best possible score with a median rank of 1, demonstrating that both the Form-Context model and EV estimate a vector that is very close to the correct one.  However, interestingly, using the subword only strategies performs better on both methods than their combined strategies (both EV-s and FCM-s are better than every other model).  This seems unusual, as we would expect context to help the subwords, and a combined representation to be better than just a subword representation.  However, this issue can be answered by looking deeper into the definitional nonce task.  The definitional nonce task tests words the model already knows (it compares the estimate to the real embedding).  Therefore, in any subword model, the subwords of the nonce words were trained with the nonce word specifically, and therefore will generally be good at estimating it.  This shows a flaw in the definitional nonce task; although it is supposed to measure how well OOV words can be estimated, it uses already learned words to do it, which may mislead on how well it performs on real OOV words.  A "better task" for this is the CRW task, which compares OOV words to other words, and therefore tests words that the models have truly never seen before.

This flaw in the definitional nonce task mainly applies to subword strategies.  As such, we also compare each context only strategy.  EV perform well using just the context, with a fairly strong rank of 74.8, compared to Form Context model's 90.7.  Interestingly,  the MRR of both context representations is reversed; the FCM-c model has a better MRR than EV-c. We suspect this means that EV context clues have a more 'stable' estimation than the FCM context estimation, where FCM context estimation is sometimes much better but can also be much worse.  MRR is affected more by individual ranks, which could explain why FCM-c has better MRR but worse median rank than EV-c.

Next, we look at the CRW task, shown in Figure \ref{fig:crw_graph}.  Like the above tasks, 10 trials of each method were analyzed, with the figure showing the average results.  Statistical significance is assessed as before, using a one-way ANOVA with a post-hoc Tukey HSD test for each context group.  All differences in performance are significant, except for fastText and FCM-s in all contexts; FCM and EV-s in later contexts (for all context sets size 8 and above); and EV with EV-s, fastText, and FCM-s in small contexts (1 and 2).

For the CRW task,  EV outperforms all other methods in when using at least 4 contexts.  In addition, it outperforms all methods in all context levels, except for fastText and FCM-s, both subword models, which it loses to in low context settings.  This demonstrates EV's strong capabilities at estimating OOV words.  As mentioned earlier, the CRW task tests a model's ability to estimate words it truly hasn't seen before, and as such we consider this task a better test of OOV word estimation.  This shows Estimator Vectors are competitive at OOV estimation.

Like the definitional nonce task, the subword only strategies perform extremely well on the CRW task, (although not as well as the full EV this time).  For the Form-Context model , the subword only model once again outperforms the combined Form-Context model, suggesting using only subwords is better than using both context and subwords with the Form-Context model.  This finding is also demonstrated by Schick and Sch{\"u}tze \cite{schick2019learning}, where subword based strategies also perform extremely well (better than all other strategies).  Schick and Sch{\"u}tze suggest this is due to the fact that CRW was built from the Rare Word dataset, which was originally contructed on words with strong morphologies.  Therefore, the rare words in this set have highly meaningful subword context, which means subword estimation strategies should do well.

When looking at context only methods, EV-c is by far the best performing method, with higher scores than FCM-c and \textit{\`{a} la carte} in any amount of contexts.  This suggests that learning a separate set of context clue embeddings (for the purpose of estimating words) seems to be an effective strategy for better OOV estimation.

Overall, the CRW task shows EV is extremely effective at estimating OOV words.

\section{Conclusion}
\label{conclusion}
	We propose Estimator Vectors (EV), a word2vec inspired model that learns high quality word embeddings, and allows for good OOV  estimates without requiring separate training. The model learns three distinct sets of embeddings: the word embedding itself, along with context clue embeddings and subword embeddings,  used for estimating OOV words.  We show this model has promising results in both word embedding quality and OOV estimation.  We plan to continue more experiments, and make our code available in the near future.
	
	We plan to extend this work.  First, we plan to experiment with various weighting strategies, in order to effectively combine context clue and subword words.  In addition, we plan to incorporate other information, like position information, to enhance the context clue representations even more.

	Furthermore, we plan to investigate ways of combining EV's word, subword, and context clue embeddings to create stronger representations for all words, not just OOV words.  With EV's context clue vectors, we can create contextualized embeddings for words using the summation of EV's various embeddings.  We plan to investigate how strong these contextualized representations can be, and how well they compare to more complex, deep contextualized representations like those generated by ELMo \cite{peters2018deep} and BERT\cite{devlin2018bert}.
    
\bibliographystyle{siamplain}
\bibliography{my_bib}

\end{document}